\documentclass[conference]{IEEEtran}

\let\labelindent\relax
\usepackage{amsmath}
\usepackage{amsfonts}
\usepackage{amssymb}
\usepackage{amsmath,amssymb,amscd,epsfig,amsfonts,rotating}
\usepackage{graphicx}
\usepackage{epstopdf}
\usepackage{subfigure}
\usepackage{multirow}
\usepackage{booktabs}
\usepackage{color,xcolor}
\usepackage{url}
\usepackage{amsmath,amscd,amsbsy,amssymb,latexsym,url,bm}
\usepackage{epsfig,epstopdf,graphicx,subfigure}
\usepackage{enumitem,balance,mathtools}
\usepackage{wrapfig}
\usepackage{mathrsfs, euscript}
\usepackage{algorithm}
\usepackage{algorithmic}
\usepackage{amssymb}
\usepackage{ifpdf}
\usepackage{diagbox}
\usepackage{caption}
\usepackage{makecell}
\usepackage{extarrows}

\newcommand{\bs}{\boldsymbol}
\newcommand{\bW}{\bs{W}}
\newcommand{\bb}{\bs{b}}
\newcommand{\bl}{\bs{l}}
\newcommand{\bz}{\bs{z}}
\newcommand{\bp}{\bs{p}}
\newcommand{\bx}{\bs{x}}
\newcommand{\Bf}{\bs{f}}
\newcommand{\btheta}{\bs{\theta}}
\newcommand{\mR}{\mathbb{R}}
\newcommand{\sigmoid}{\text{sigmoid}}
\newcommand{\relu}{\text{relu}}

\begin{document}

\title{Product-based Neural Networks for User Response Prediction}

\author{
\IEEEauthorblockN{Yanru Qu, Han Cai, Kan Ren, Weinan Zhang, Yong Yu}
\IEEEauthorblockA{Shanghai Jiao Tong University\\
	\{kevinqu, hcai, kren, wnzhang, yyu\}@apex.sjtu.edu.cn}
\and
\IEEEauthorblockN{Ying Wen, Jun Wang}
\IEEEauthorblockA{University College London\\
\{ying.wen, j.wang\}@cs.ucl.ac.uk}
}

\maketitle

\begin{abstract}
Predicting user responses, such as clicks and conversions, is of great importance and has found its usage in many Web applications including recommender systems, web search and online advertising. The data in those applications is mostly categorical and contains multiple fields; a typical representation is to transform it into a high-dimensional sparse binary feature representation via one-hot encoding.
Facing with the extreme sparsity, traditional models may limit their capacity of mining shallow patterns from the data, i.e. low-order feature combinations. Deep models like deep neural networks, on the other hand, cannot be directly applied for the high-dimensional input because of the huge feature space.
In this paper, we propose a Product-based Neural Networks (PNN) with an embedding layer to learn a distributed representation of the categorical data, a product layer to capture interactive patterns between inter-field categories, and further fully connected layers to explore high-order feature interactions.
Our experimental results on two large-scale real-world ad click datasets demonstrate that PNNs consistently outperform the state-of-the-art models on various metrics.
\end{abstract}




\IEEEpeerreviewmaketitle

\section{Introduction}\label{sec:intro}
Learning and predicting user response now plays a crucial role in many personalization tasks in information retrieval (IR), such as recommender systems, web search and online advertising. The goal of user response prediction is to estimate the probability that the user will provide a predefined positive response, e.g. clicks, purchases etc., in a given context
\cite{menon2011response}.
This predicted probability indicates the user's interest on the specific item such as a news article, a commercial item or an advertising post, which influences the subsequent decision making such as document ranking \cite{xue2004optimizing} and ad bidding \cite{zhang2014optimal}.


The data collection in these IR tasks is mostly in a multi-field categorical form, for example, \texttt{[Weekday=Tuesday, Gender=Male, City=London]}, which is normally transformed into high-dimensional sparse binary features via one-hot encoding \cite{he2014practical}.
For example, the three field vectors with one-hot encoding are concatenated as
\[ \underbrace{[0,1,0,0,0,0,0]}_{\texttt{Weekday=Tuesday}}\underbrace{[0,1]}_{\texttt{Gender=Male}} \underbrace{[0,0,1,0,\ldots,0,0]}_{\texttt{City=London}}.\]
Many machine learning models, including linear logistic regression \cite{lee2012estimating}, non-linear gradient boosting decision trees \cite{he2014practical} and factorization machines \cite{ta2015factorization}, have been proposed to work on such high-dimensional sparse binary features and produce high quality user response predictions.
However, these models highly depend on feature engineering in order to capture high-order latent patterns \cite{cui2011bid}.

Recently, deep neural networks (DNNs) \cite{lecun2015deep} have shown great capability in classification and regression tasks, including computer vision \cite{krizhevsky2012imagenet}, speech recognition \cite{graves2013speech} and natural language processing \cite{mikolov2013distributed}. It is promising to adopt DNNs in user response prediction since DNNs could automatically learn more expressive feature representations and deliver better prediction performance.
In order to improve the multi-field categorical data interaction, \cite{zhang2016deep} presented an embedding methodology based on pre-training of a factorization machine. Based on the concatenated embedding vectors, multi-layer perceptrons (MLPs) were built to explore feature interactions. However, the quality of embedding initialization is largely limited by the factorization machine.
More importantly, the ``add'' operations of the perceptron layer might not be useful to explore the interactions of categorical data in multiple fields. Previous work \cite{menon2011response,ta2015factorization} has shown that local dependencies between features from different fields can be effectively explored by feature vector ``product'' operations instead of ``add'' operations.



To utilize the learning ability of neural networks and mine the latent patterns of data in a more effective way than \text{MLPs,} in this paper we propose Product-based Neural Network (PNN) which (i) starts from an embedding layer without pre-training as used in \cite{zhang2016deep}, and (ii) builds a product layer based on the embedded feature vectors to model the inter-field feature interactions, and (iii) further distills the high-order feature patterns with fully connected MLPs.
We present two types of PNNs, with inner and outer product operations in the product layer, to efficiently model the interactive patterns.

We take CTR estimation in online advertising as the working example to explore the learning ability of our PNN model. The extensive experimental results on two large-scale real-world datasets demonstrate the consistent superiority of our model over state-of-the-art user response prediction models \cite{ta2015factorization,liu2015convolutional,zhang2016deep} on various metrics.

\section{Related Work}\label{sec:related}
The response prediction problem is normally formulated as a binary classification problem with prediction likelihood or cross entropy as the training objective \cite{richardson2007predicting}.
Area under ROC Curve (AUC) and Relative Information Gain (RIG) are common evaluation metrics for response prediction accuracy \cite{graepel2010web}.
From the modeling perspective, linear logistic regression (LR) \cite{lee2012estimating,ren2016user} and non-linear gradient boosting decision trees (GBDT) \cite{he2014practical} and factorization machines (FM) \cite{ta2015factorization} are widely used in industrial applications. However, these models are limited in mining high-order latent patterns or learning quality feature representations.

Deep learning is able to explore high-order latent patterns as well as generalizing expressive data representations \cite{mikolov2013distributed}.
The input data of DNNs are usually dense real vectors, while the solution of multi-field categorical data has not been well studied. Factorization-machine supported neural networks (FNN) was proposed in \cite{zhang2016deep} to learn embedding vectors of categorical data via pre-trained FM.
Convolutional Click Prediction Model (CCPM) was proposed in \cite{liu2015convolutional} to predict ad click by convolutional neural networks (CNN). However, in CCPM the convolutions are only performed on the neighbor fields in a certain alignment, which fails to model the full interactions among non-neighbor features.
Recurrent neural networks (RNN) was leveraged to model the user queries as a series of user context to predict the ad click behavior \cite{zhang2014sequential}.
Product unit neural network (PUNN) \cite{engelbrecht1999training} was proposed to build high-order combinations of the inputs. However, neither can PUNN learn local dependencies, nor produce bounded outputs to fit the response rate.

In this paper, we demonstrate the way our PNN models learn local dependencies and high-order feature interactions.

\section{Deep Learning for CTR Estimation}\label{sec:methodology}

We take CTR estimation in online advertising \cite{richardson2007predicting} as a working example to formulate our model and explore the performance on various metrics. The task is to build a prediction model to estimate the probability of a user clicking a specific ad in a given context.

Each data sample consists of multiple fields of categorical data such as user information (\texttt{City}, \texttt{Hour}, etc.), publisher information (\texttt{Domain}, \texttt{Ad slot}, etc.) and ad information (\texttt{Ad creative ID}, \texttt{Campaign ID}, etc.) \cite{zhang2014real}. All the information is represented as a multi-field categorical feature vector, where each field (e.g. \texttt{City}) is one-hot encoded as discussed in Section~\ref{sec:intro}.
Such a field-wise one-hot encoding representation results in curse of dimensionality and enormous sparsity \cite{zhang2016deep}. Besides, there exist local dependencies and hierarchical structures among fields \cite{menon2011response}.

Thus we are seeking a DNN model to capture high-order latent patterns in multi-field categorical data. And we come up with the idea of product layers to explore feature interactions automatically. In FM, feature interaction is defined as the inner product of two feature vectors \cite{rendle2010factorization}.

The proposed deep learning model is named as Product-based Neural Network (PNN). In this section, we present PNN model in detail and discuss two variants of this model, namely Inner Product-based Neural Network (IPNN), which has an inner product layer, and Outer Product-based Neural Network (OPNN) which uses an outer product expression.



\subsection{Product-based Neural Network}



\begin{figure}[t]
	\centering
	\includegraphics[width=1\columnwidth]{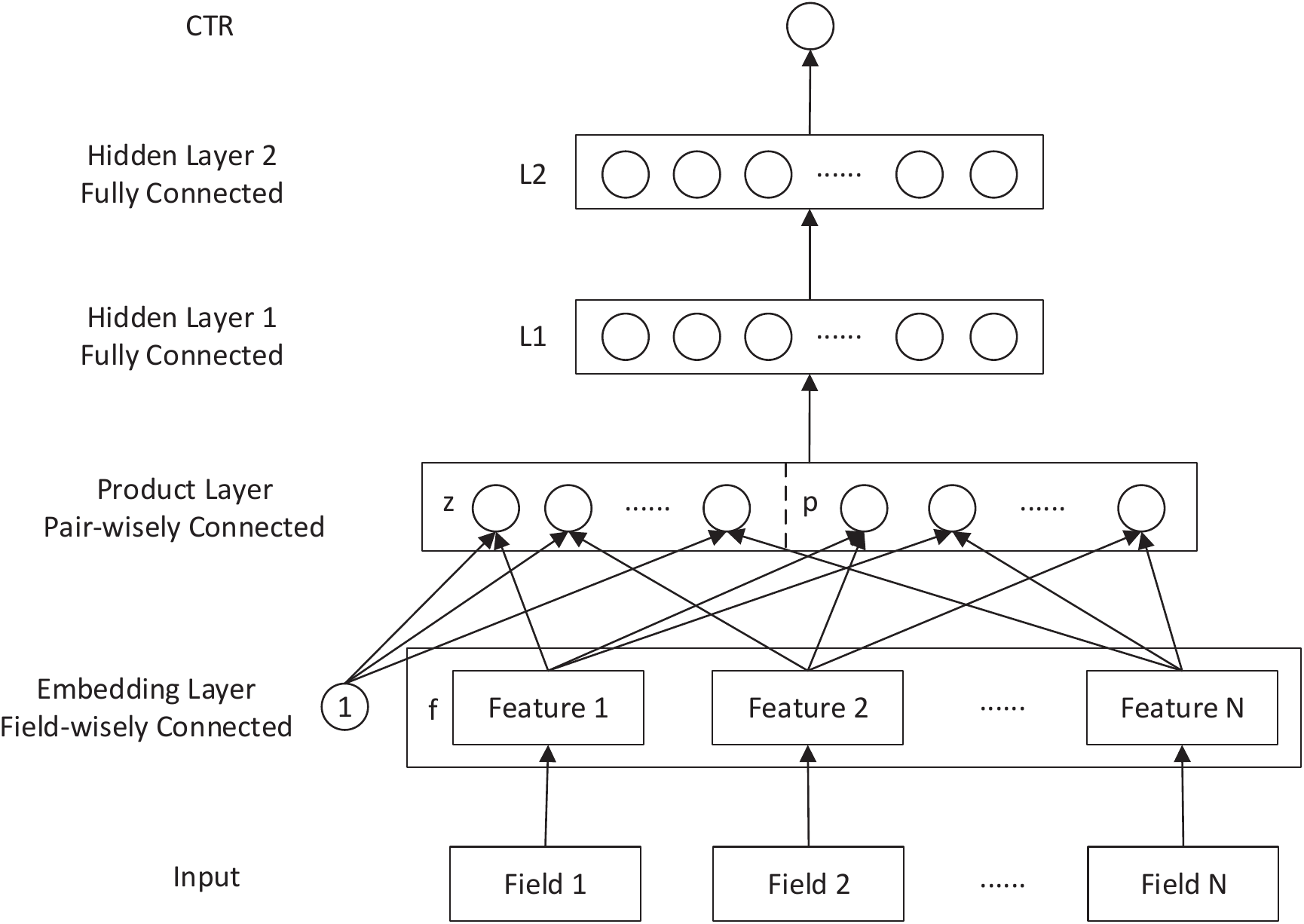}
	\caption{Product-based Neural Network Architecture.}\label{fig:pnn}
\end{figure}

The architecture of the PNN model is illustrated in Figure~\ref{fig:pnn}.
From a top-down perspective, the output of PNN is a real number $\hat{y} \in ( 0, 1 )$ as the predicted CTR:
\begin{equation}
\hat{y} = \sigma(\bW_3 \bl_2 + b_3), \label{eq:output}
\end{equation}
where $\bW_3 \in \mR^{1 \times D_2}$ and $b_3 \in \mR$ are the parameters of the output layer, $\bl_2 \in \mR^{D_2}$ is the output of the second hidden layer, and $\sigma(x)$ is the sigmoid activation function: $\sigma(x)= 1 / (1 + e^{-x})$. And we use $D_i$ to represent the dimension of the $i$-th hidden layer.

The output $\bl_2$ of the second hidden layer is constructed as
\begin{equation}
\bl_2 = \relu(\bW_2 \bl_1 + \bb_2),
\end{equation}
where $\bl_1 \in \mR^{D_1}$ is the output of the first hidden layer. The rectified linear unit (relu), defined as $\relu(x) = \max(0, x)$, is chosen as the activation function for hidden layer output since it has outstanding performance and efficient computation.

The first hidden layer is fully connected with the product layer. The inputs to it consist of linear signals $\bl_z$ and quadratic signals $\bl_p$. With respect to $\bl_z$ and $\bl_p$ inputs, separately, the formulation of $\bl_1$ is:
\begin{align}\label{eq:l1-formula}
\bl_1 &= \relu(\bl_z + \bl_p + \bb_1),
\end{align}
where all $\bl_z$, $\bl_p$ and the bias vector $\bb_1 \in \mR^{D_1}$.

Then, let us define the operation of tensor inner product:
\begin{align}
\bs{A} & \odot \bs{B} \triangleq \sum_{i,j} \bs{A}_{i,j} \bs{B}_{i,j},
\end{align}
where firstly element-wise multiplication is applied to $\bs{A}, \bs{B}$, then the multiplication result is summed up to a scalar. After that,
$\bl_z$ and $\bl_p$ are calculated through $\bz$ and $\bp$, respectively:
\begin{align}\label{eq:l1-lz-lp}
\bl_z &= \begin{pmatrix}
l_z^1, l_z^2, \ldots, l_z^n, \ldots , l_z^{D_1}
\end{pmatrix}, \qquad l_z^n = \bW_z^n \odot \bz\\
\bl_p &= \begin{pmatrix}
l_p^1, l_p^2, \ldots, l_p^n, \ldots , l_p^{D_1}
\end{pmatrix}, \qquad l_p^n = \bW_p^n \odot \bp  \nonumber
\end{align}
where $\bW_z^n$ and $\bW_p^n$ are the weights in the product layer, and their shapes are determined by $\bz$ and $\bp$ respectively.

By introducing a ``1" constant signal, the product layer can not only generate the quadratic signals $\bp$, but also maintaining the linear signals $\bz$, as illustrated in Figure \ref{fig:pnn}. More specifically,
\begin{align}
\bz &= \begin{pmatrix}
\bz_1, \bz_2, \ldots, \bz_N
\end{pmatrix}
\triangleq \begin{pmatrix}
\Bf_1, \Bf_2, \ldots, \Bf_N
\end{pmatrix}, \\
\bp	&=
\{
\bp_{i, j}
\}, i = 1...N, j = 1...N,
\end{align}
where $\Bf_i \in \mathbb{R}^M$ is the embedding vector for field $i$. $\bp_{i, j} = g(\Bf_i, \Bf_j)$ defines the pairwise feature interaction.
Our PNN model can have different implementations by designing different operation for $g$. In this paper, we propose two variants of PNN, namely IPNN and OPNN, as will be discussed later.

The embedding vector $\Bf_i$ of field $i$, is the output of the embedding layer:
\begin{equation}
\Bf_i = \bW_0^i~\bx[\text{start}_i : \text{end}_i],
\end{equation}
where $\bx$ is the input feature vector containing multiple fields, and $\bx[\text{start}_i : \text{end}_i]$ represents the one-hot encoded vector for field $i$. $\bW_0$ represents the parameters of the embedding layer, and $\bW_0^i \in \mR^{M \times (\text{end}_i - \text{start}_i + 1)}$ is fully connected with field $i$.

Finally, supervised training is applied to minimize the log loss, which is a widely used objective function capturing divergence between two probability distributions: 
\begin{equation}\label{log_loss}
L(y,\hat{y}) = -y \log\hat{y} - (1-y) \log(1-\hat{y}),
\end{equation}
where $y$ is the ground truth (1 for click, 0 for non-click), and $\hat{y}$ is the predicted CTR of our model as in Eq.~(\ref{eq:output}).

\subsection{Inner Product-based Neural Network}\label{sec:pnn-i}
In this section, we demonstrate the Inner Product-based Neural Network (IPNN). In IPNN, we firstly define the pairwise feature interaction as vector inner product
: $g(\Bf_i, \Bf_j) = \langle \Bf_i,\Bf_j \rangle$.



With the constant signal ``1", the linear information $\bz$ is preserved as:
\begin{equation}\label{eq:lz}
\bl_z^n = \bW_z^n \odot \bz = \sum_{i=1}^{N} {\sum_{j=1}^{M} {(\bW_z^n)_{i,j} \bz_{i,j}}}.
\end{equation}

As for the quadratic information $\bp$, the pairwise inner product terms of $g(\Bf_i, \Bf_j)$ form a square matrix $\bp \in \mR^{N \times N}$.
Recalling the definition of $\bl_p$ in Eq.~(\ref{eq:l1-lz-lp}), $l_p^n = \sum_{i = 1}^{N} \sum_{j = 1}^{N} (\bW_p^n)_{i, j} \bp_{i, j}$ and the commutative law in vector inner product, $\bp$ and $\bW_p^n$ should be symmetric.

Such pairwise connection expands the capacity of the neural network, but also enormously increases the complexity.
In this case, the formulation of $\bl_1$, described in Eq.~(\ref{eq:l1-formula}), has the space complexity of $O(D_1 N (M + N))$, and the time complexity of $O(N^2(D_1 + M))$, where $D_1$ and $M$ are the hyper-parameters about network architecture, $N$ is the number of input fields.
Inspired by FM \cite{rendle2010factorization}, we come up with the idea of matrix factorization to reduce complexity.



By introducing the assumption that $\bW_p^n = \btheta^n {\btheta^n}^T$, where $ \btheta^n \in \mR^N$, we can simplify $\bl_1$'s formulation as:
\begin{equation}
\bW_p^n \odot \bp = \sum_{i=1}^N {\sum_{j=1}^N \theta^n_i \theta^n_j \langle \Bf_i, \Bf_j \rangle} = \langle \sum_{i=1}^N\bs\delta^n_i, \sum_{i=1}^N\bs\delta^n_i \rangle
\end{equation}
where, for convenience, we use $\bs\delta^n_i \in \mR^M$ to denote a feature vector $\Bf_i$ weighted by $\theta^n_i$, i.e. $\bs\delta^n_i = \theta^n_i \Bf_i$. And we also have $\bs\delta^n = \begin{pmatrix} \bs\delta^n_1, \bs\delta^n_2, \ldots, \bs\delta^n_i, \ldots, \bs\delta^n_N \end{pmatrix} \in \mR^{N \times M}$.

With the first order decomposition on $n$-th single node, we give the $\bl_p$ complete form:

\begin{equation}\label{eq:pnn1_lp}
\begin{split}
\bl_p = \Big( \Vert \sum_i \bs\delta_i^1 \Vert, \ldots, \Vert \sum_i \bs\delta_i^{n} \Vert, \ldots, \Vert \sum_i \bs\delta_i^{D_1} \Vert \Big).
\end{split}
\end{equation}


By reduction of $\bl_p$ in Eq.~(\ref{eq:pnn1_lp}), the space complexity of $\bl_1$ becomes $O(D_1MN)$, and the time complexity is also $O(D_1MN)$. In general, $\bl_1$ complexity is reduced from quadratic to linear with respect to $N$. This well-formed equation makes reusable for some intermediate results. Moreover, matrix operations are easily accelerated in practice with GPUs.

More generally, we discuss $K$-order decomposition of $\bW_p^n$ at the end of this section. We should point out that $\bW_p^n = \bs\theta_n \bs\theta_n^T$ is only the first order decomposition with a strong assumption. The general matrix decomposition method can be derived that:
\begin{equation}
\bW_p^n \odot \bp = \sum_{i=1}^N {\sum_{j=1}^N \langle \bs\theta_n^i, \bs\theta_n^j\rangle \langle \Bf_i, \Bf_j \rangle}.
\end{equation}
In this case, $\bs\theta_n^i \in \mR^K$. This general decomposition is more expressive with weaker assumptions, but also leading to $K$ times model complexity.

\subsection{Outer Product-based Neural Network}\label{sec:pnn-ii}
Vector inner product takes a pair of vectors as input and outputs a scalar. Different from that, vector outer product takes a pair of vectors and produces a matrix. IPNN defines feature interaction by vector inner product, while in this section, we discuss the Outer Product-based Neural Network (OPNN).

The only difference between IPNN and OPNN is the quadratic term $\bp$.
In OPNN, we define feature interaction as $g(\Bf_i,\Bf_j) = \Bf_i \Bf_j^T$. Thus for every element in $\bp$, $\bp_{i,j} \in \mR^{M \times M}$ is a square matrix.

For calculating $\bl_1$, the space complexity is $O(D_1M^2N^2)$ , and the time complexity is also $O(D_1M^2N^2)$. Recall that $D_1$ and $M$ are the hyper-parameters of the network architecture, and $N$ is the number of the input fields, this implementation is expensive in practice. To reduce the complexity, we propose the idea of \emph{superposition}.

By element-wise superposition, we can reduce the complexity by a large step. Specifically, we re-define $\bp$ formulation as
\begin{equation}
\bp = \sum_{i=1}^{N}{\sum_{j=1}^N{\Bf_i \Bf_j^T}} = \Bf_{\Sigma} (\Bf_{\Sigma})^T, \quad \Bf_{\Sigma} = \sum_{i=1}^N {\Bf_i},
\end{equation}
where $\bp \in \mR^{M \times M}$ becomes symmetric, thus $\bW_p^n$ should also be symmetric.
Recall Eq.~(\ref{eq:l1-lz-lp}) that $\bW_p \in \mR^{D_1 \times M \times M}$. In this case, the space complexity of $\bl_1$ becomes $O(D_1M(M+N))$, and the time complexity is also $O(D_1M(M+N))$.

%


\subsection{Discussions}\label{sec:general-pnn}

Compared with FNN \cite{zhang2016deep}, PNN has a product layer. If removing $\bl_{\bp}$ part of the product layer, PNN is identical to FNN. With the inner product operator, PNN is quite similar with FM \cite{rendle2010factorization}: if there is no hidden layer and the output layer is simply summing up with uniform weight, PNN is identical to FM. Inspired by Net2Net \cite{chen2015net2net}, we can firstly train a part of PNN (e.g., the FNN or FM part) as the initialization, and then start to let the back propagation go over the whole net. The resulted PNN should at least be as good as FNN or FM.

In general, PNN uses product layers to explore feature interactions. Vector products can be viewed as a series of addition/multiplication operations. Inner product and outer product are just two implementations. In fact, we can define more general or complicated product layers, gaining PNN better capability in exploration of feature interactions.

Analogous to electronic circuit, addition acts like ``OR" gate while multiplication acting like ``AND" gate, and the product layer seems to learn rules other than features. Reviewing the scenario of computer vision, while pixels in images are real-world raw features, categorical data in web applications are artificial features with high levels and rich meanings. Logic is a powerful tool in dealing with concepts, domains and relationships. Thus we believe that introducing product operations in neural networks will improve networks' ability for modeling multi-field categorical data.

\section{Experiments}\label{sec:experiment}
In this section, we present our experiments in detail, including datasets, data processing, experimental setup, model comparison, and the corresponding analysis\footnote{We release the repeatable experiment code on GitHub: \url{https://github.com/Atomu2014/product-nets}}. In our experiments, PNN models outperform major state-of-the-art models in the CTR estimation task on two real-world datasets.

\subsection{Datasets}
\subsubsection{Criteo}
Criteo 1TB click log\footnote{Criteo terabyte dataset download link: \url{http://labs.criteo.com/downloads/download-terabyte-click-logs/}.} is a famous ad tech industry benchmarking dataset.
We select 7 consecutive days of samples for training, and the next 1 day for evaluation. Because of the enormous data volume and high bias, we apply negative down-sampling on this dataset.
Define the down-sampling ratio as $w$, the predicted CTR as $p$, the re-calibrated CTR $q$ should be $q =p / (p + \frac{1-p}{w})$ \cite{he2014practical}.
After down-sampling and feature mapping, we get a dataset, which comprises 79.38M instances with 1.64M feature dimensions.

\subsubsection{iPinYou}
The iPinYou dataset\footnote{iPinYou dataset download link: \url{http://data.computational-advertising.org}. We only use the data from season 2 and 3 because of the same data schema.} is another real-world dataset for ad click logs over 10 days.
After one-hot encoding, we get a dataset containing 19.50M instances with 937.67K input dimensions.
We keep the original train/test splitting scheme, where for each advertiser the last 3-day data are used as the test dataset while the rest as the training dataset.


\subsection{Model Comparison}
We compare 7 models in our experiments, which are implemented with TensorFlow\footnote{TensorFlow: \url{https://www.tensorflow.org/}}, and trained with Stochastic Gradient Descent (SGD).

\textbf{LR}: LR is the most widely used linear model in industrial applications \cite{mcmahan2013ad}. It is easy to implement and fast to train, however, unable to capture non-linear information.

\textbf{FM}: FM has many successful applications in recommender systems and user response prediction tasks \cite{rendle2010factorization}. FM explores feature interactions, which is effective on sparse data.

\textbf{FNN}: FNN is proposed in \cite{zhang2016deep}, being able to capture high-order latent patterns of multi-field categorical data.

\textbf{CCPM}: CCPM is a convolutional model for click prediction \cite{liu2015convolutional}. This model learns local-global features efficiently. However, CCPM highly relies on feature alignment, and is lack of interpretation.

\textbf{IPNN}: PNN with inner product layer \ref{sec:pnn-i}.

\textbf{OPNN}: PNN with outer product layer \ref{sec:pnn-ii}.

\textbf{PNN*}: This model has a product layer, which is a concatenation of inner product and outer product.



Additionally, in order to prevent over-fitting, the popular L2 regularization term is added to the loss function $ L(y,\hat{y})$ when training LR and FM.
And we also employ dropout as a regularization method to prevent over-fitting when training neural networks.


\subsection{Evaluation Metrics}
Four evaluation metrics are tested in our experiments. The two major metrics are:

\textbf{AUC}: Area under ROC curve is a widely used metric in evaluating classification problems. Besides, some work validates AUC as a good measurement in CTR estimation \cite{graepel2010web}.

\textbf{RIG}: Relative Information Gain, $RIG = 1 - NE$, where NE is the Normalized Cross Entropy \cite{he2014practical}.

Besides, we also employ \textbf{Log Loss} (Eq.~(\ref{log_loss})) and root mean square error (\textbf{RMSE}) as our additional evaluation metrics.

\subsection{Performance Comparison}

\begin{table}[t]
	\centering
	\caption{Overall Performance on the Criteo Dataset.}\label{tab:perf-criteo}
	\begin{tabular}{m{40pt}<{\centering} | m{35pt}<{\centering} m{35pt}<{\centering} m{35pt}<{\centering} m{35pt}<{\centering}}
		\hline
		Model & AUC & Log Loss & RMSE & RIG \\
		\hline
		LR & 71.48\% & 0.1334 & 9.362e-4 & 6.680e-2 \\
		FM & 72.20\% & 0.1324 & 9.284e-4 & 7.436e-2 \\
		FNN & 75.66\% & 0.1283 & 9.030e-4 & 1.024e-1 \\
		CCPM & 76.71\% & 0.1269 & 8.938e-4 & 1.124e-1 \\
		IPNN & \textbf{77.79\%} & \textbf{0.1252} & \textbf{8.803e-4} & \textbf{1.243e-1} \\
		OPNN & 77.54\% & 0.1257 & 8.846e-4 & 1.211e-1 \\
		PNN* & 77.00\% & 0.1270 & 8.988e-4 & 1.118e-1\\
		\hline
	\end{tabular}
\vspace{10pt}
	\centering
	\caption{Overall Performance on the iPinYou Dataset.}\label{tab:perf-ipinyou}
	\begin{tabular}{m{40pt}<{\centering} | m{35pt}<{\centering} m{35pt}<{\centering} m{35pt}<{\centering} m{35pt}<{\centering}}
		\hline
		Model & AUC & Log Loss & RMSE & RIG \\
		\hline
		LR & 73.43\% & 5.581e-3 & 5.350e-07 & 7.353e-2 \\
		FM & 75.52\% & 5.504e-3 & 5.343e-07 & 8.635e-2 \\
		FNN & 76.19\% & 5.443e-3 & 5.285e-07 & 9.635e-2 \\
		CCPM & 76.38\% & 5.522e-3 & 5.343e-07 & 8.335e-2 \\
		IPNN & 79.14\% & 5.195e-3 & 4.851e-07 & 1.376e-1\\
		OPNN & \textbf{81.74\%} & 5.211e-3 & 5.293e-07 & 1.349e-1 \\
		PNN* & 76.61\% & \textbf{4.975e-3} & \textbf{4.819e-07} & \textbf{1.740e-1}\\
		\hline
	\end{tabular}
	\centering
	\vspace{10pt}
    \caption{P-values under the Log Loss Metric.}\label{tab:pvalue}
	\begin{tabular}{m{40pt}<{\centering} | m{35pt}<{\centering} m{35pt}<{\centering} m{35pt}<{\centering} m{35pt}<{\centering}}
		\hline
		Model & LR & FM & FNN & CCPM \\
		\hline
		IPNN & $< 10^{-6}$ & $< 10^{-6}$ & $< 10^{-6}$ & $< 10^{-6}$ \\
		OPNN & $< 10^{-6}$ & $< 10^{-5}$ & $< 10^{-6}$ & $< 10^{-6}$ \\
		\hline
	\end{tabular}
\end{table}

Table~\ref{tab:perf-criteo} and \ref{tab:perf-ipinyou} show the overall performance on Criteo and iPinYou datasets, respectively.
In FM, we employ 10-order factorization and correspondingly, we employ 10-order embedding in network models. CCPM has 1 embedding layer, 2 convolution layers (with max pooling) and 1 hidden layer (5 layers in total). FNN has 1 embedding layer and 3 hidden layers (4 layers in total).
Every PNN has 1 embedding layer, 1 product layer and 3 hidden layers (5 layers in total). The impact of network depth will be discussed later.

\begin{figure}[t]
	\centering
	\includegraphics[width=0.7\columnwidth]{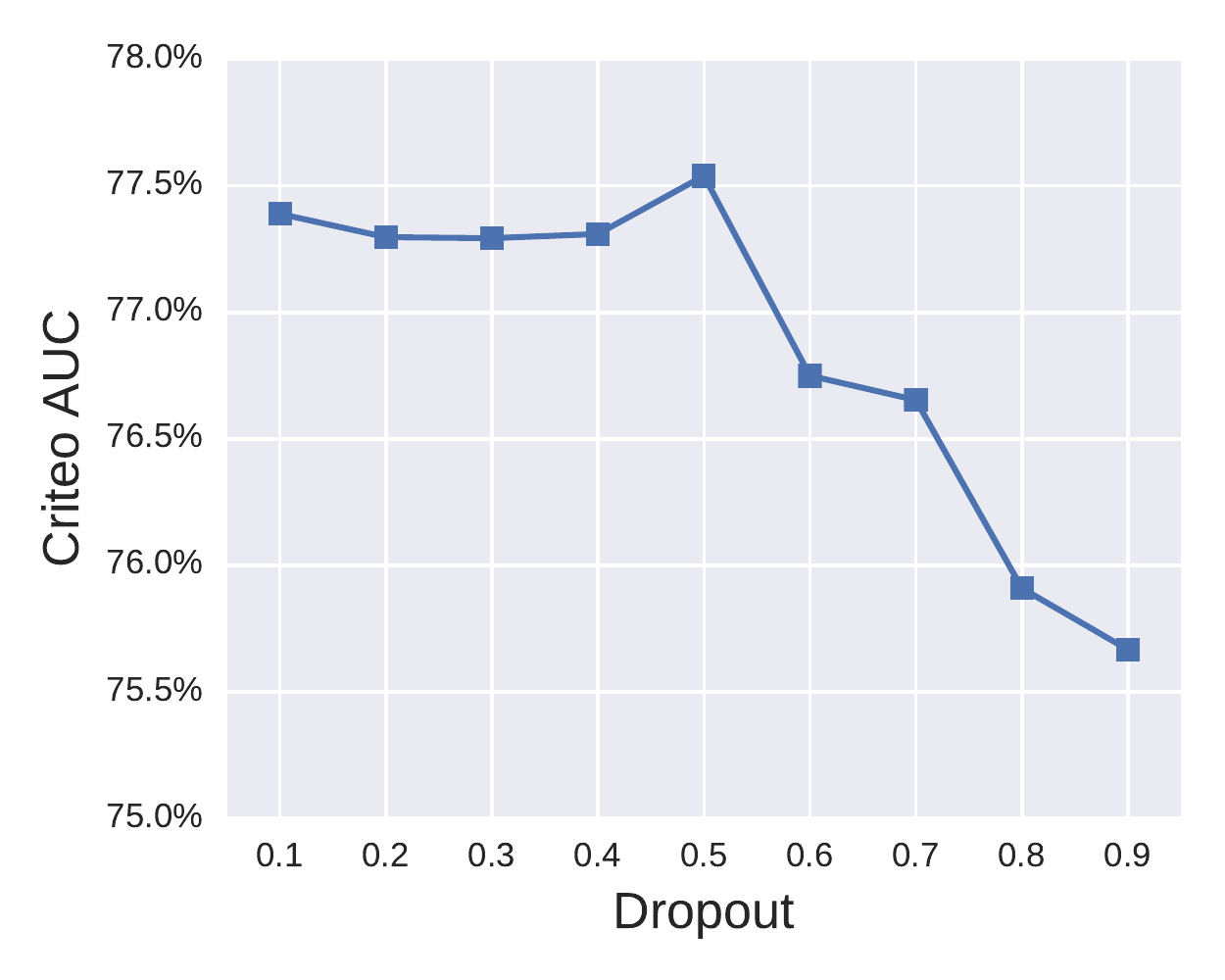}
\vspace{-5pt}	
\caption{AUC Comparison of Dropout (OPNN).}\label{fig:drop-auc}
\vspace{-5pt}	
\end{figure}

The LR and FM models are trained with L2 norm regularization, while FNN, CCPM and PNNs are trained with dropout.
By default, we set dropout rate at 0.5 on network hidden layers, which is proved effective in Figure~\ref{fig:drop-auc}.
Further discussions about the network architecture will be provided in Section~\ref{sec:arch}.

Firstly, we focus on the AUC performance. The overall results in Table \ref{tab:perf-criteo} and \ref{tab:perf-ipinyou} illustrate that (i) FM outperforms LR, demonstrating the effectiveness of feature interactions; (ii) Neural networks outperform LR and FM, which validates the importance of high-order latent patterns; (iii) PNNs perform the best on both Criteo and iPinYou datasets. As for log loss, RMSE and RIG, the results are similar.

We also conduct t-test between our proposed PNNs and the other compared models. Table~\ref{tab:pvalue} shows the calculated p-values under log loss metric on both datasets. The results verify that our models significantly improve the performance of user response prediction against the baseline models.





We also find that PNN*, which is the combination of IPNN and OPNN, has no obvious advantages over IPNN and OPNN on AUC performance.
We consider that IPNN and OPNN are sufficient to capture the feature interactions in multi-field categorical data.




\begin{figure}[t]
	\centering
	\includegraphics[width=0.7\columnwidth]{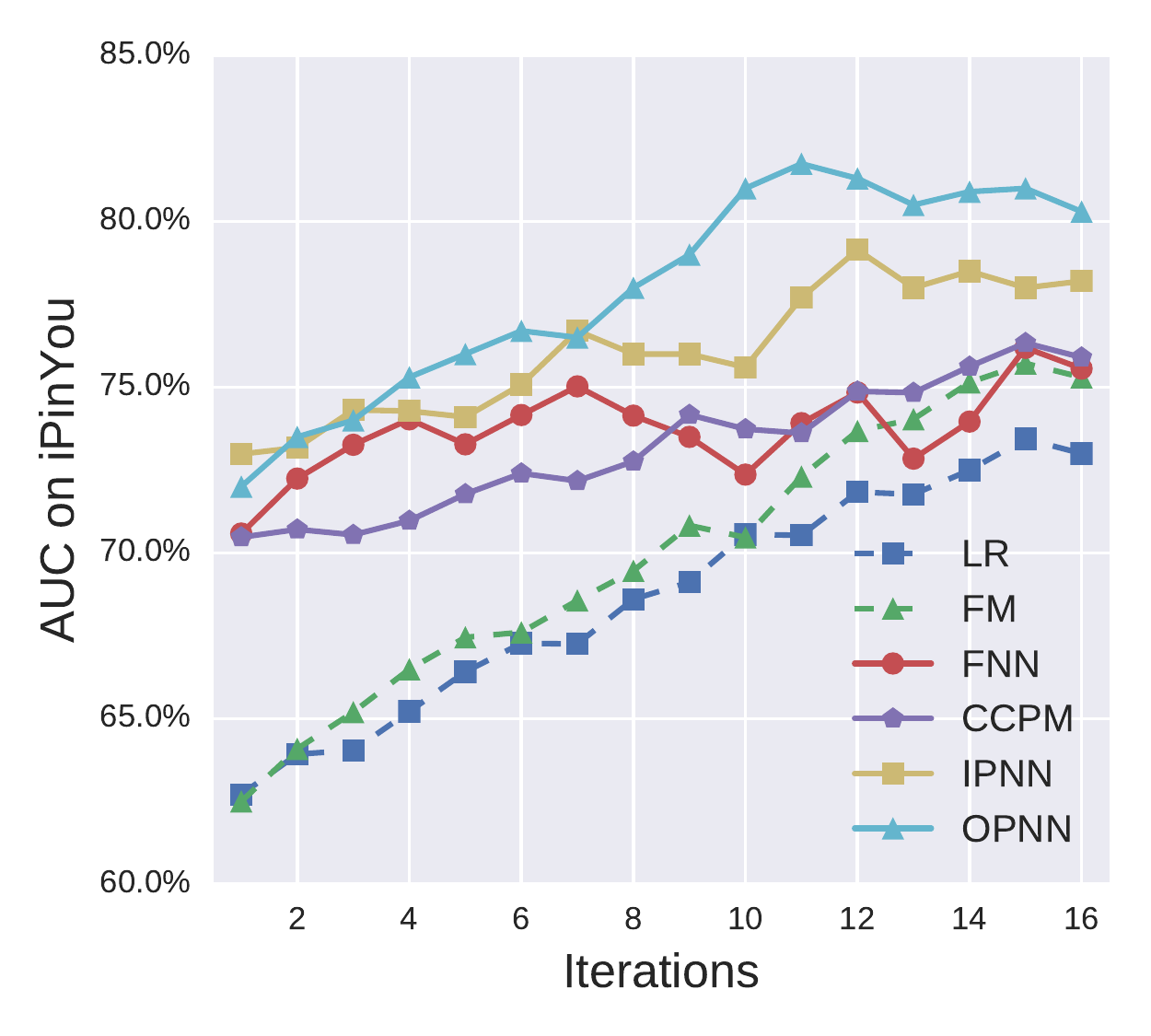}
	\caption{Learning Curves on the iPinYou Dataset.}\label{fig:train}
\end{figure}

Figure \ref{fig:train} shows the AUC performance with respect to the training iterations on iPinYou dataset. We find that network models converge more quickly than LR and FM. We also observe that our two proposed PNNs have better convergence than other network models.

\subsection{Ablation Study on Network Architecture}\label{sec:arch}
In this section, we discuss the impact of neural network architecture.
For IPNN and OPNN,
we take three hyper-parameters (or settings) into consideration:
(i) embedding layer size, (ii) network depth and (iii) activation function. Since CCPM shares few similarities with other neural networks and PNN* is just a combination of IPNN and OPNN, we only compare FNN, IPNN and OPNN in this section.

\subsubsection{Embedding Layer}
The embedding layer is to convert sparse binary inputs to dense real-value vectors.
Take word embedding as an example \cite{mikolov2013distributed},
an embedding vector contains the information of the word and its context, and indicates the relationships between words.

We take the idea of embedding layer from \cite{zhang2016deep}. In this paper, the latent vectors learned by FM are explained as node representations, and the authors use a pre-trained FM to initialize the embedding layers in FNN. Thus the factorization order of FM keeps consistent with the embedding order.

The input units are fully connected with the embedding layer within each field.
We compare different orders, like 2, 10, 50 and 100. However, when the order grows larger, it is harder to fit the parameters in memory, and the models are much easier to over-fit.
In our experiments, we take 10-order embedding in neural networks.


\subsubsection{Network Depth}

\begin{figure}[t]
	\centering
	\includegraphics[width=0.5\textwidth]{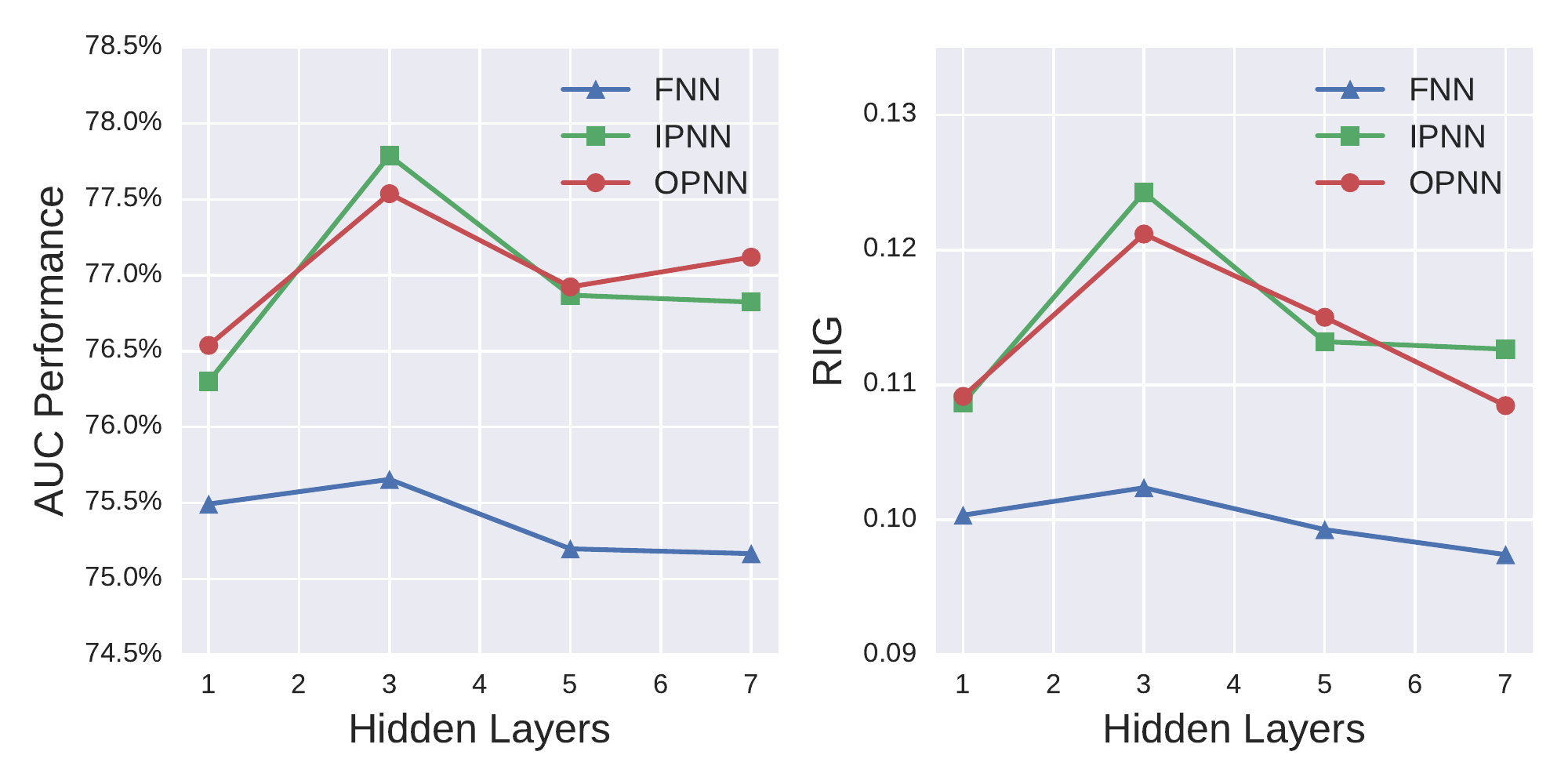}
	\caption{Performance Comparison over Network Depths.}\label{fig:depth}
\end{figure}

We also explore the impact of network depth by adjusting the number of hidden layers in FNN and PNNs.
We compare different number of hidden layers: 1, 3, 5 and 7. Figure \ref{fig:depth} shows the performance as network depth grows. Generally speaking, the networks with 3 hidden layers have better generalization on the test set.

For convenience, we call convolution layers and product layers as representation layers.
These layers can capture complex feature patterns using fewer parameters, thus are efficient in training, and generalize better on the test set.


\subsubsection{Activation Function}
We compare three mainstream activation functions: $\sigmoid(x) = \frac{1}{1+e^{-x}}$, $\tanh(x) = \frac{1 - e^{-2x}}{1 + e^{-2x}}$, and $\relu(x) = \max(0, x)$.
Compared with the sigmoidal family, relu function has the advantages of sparsity and efficient gradient, which is possible to gain more benefits on multi-field categorical data.

\begin{figure}[t]
	\centering
	\includegraphics[width=0.5\textwidth]{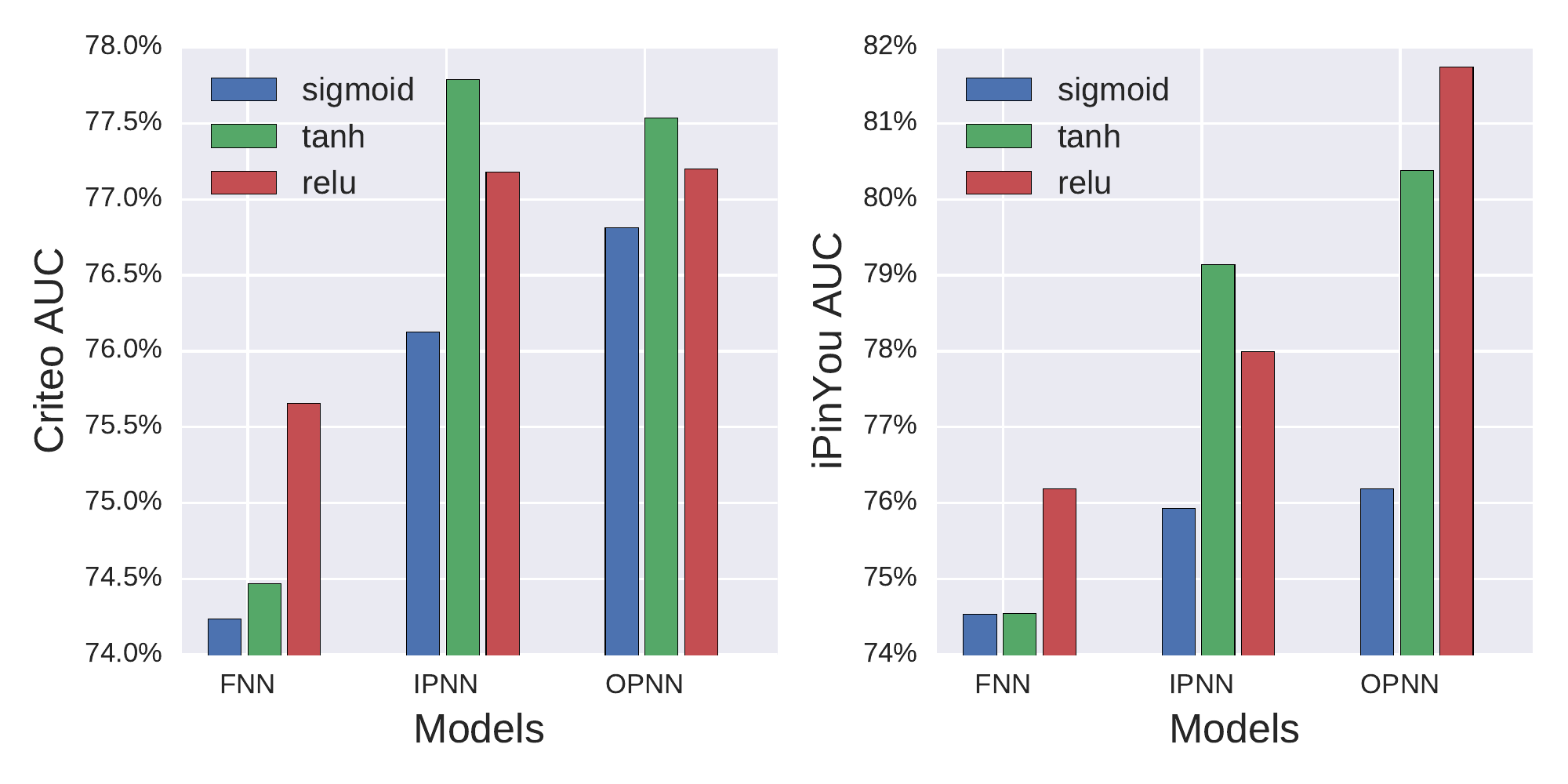}
	\caption{AUC Comparison over Various Activation Functions.}\label{fig:act}
\end{figure}

Figure \ref{fig:act} compares these activation functions on FNN, IPNN and OPNN. From this figure, we find that tanh has better performance than sigmoid. This is supported by \cite{zhang2016deep}.
Besides tanh, we find relu function also has good performance. Possible reasons include: (i) Sparse activation, nodes with negative outputs will not be activated; (ii) Efficient gradient propagation, no vanishing gradient problem or exploding effect; (iii) Efficient computation, only comparison, addition and multiplication.

\section{Conclusion and Future Work}\label{sec:conclusion}
In this paper, we proposed a deep neural network model with novel architecture, namely Product-based Neural Network, to improve the prediction performance of DNN working on categorical data. And we chose CTR estimation as our working example. By exploration of feature interactions, PNN
is promising to learn high-order latent patterns on multi-field categorical data.
We designed two types of PNN: IPNN based on inner product and OPNN based on outer product. We also discussed solutions to reduce complexity, making PNN efficient and scalable. Our experimental results demonstrated that PNN outperformed the other state-of-the-art models in 4 metrics on 2 datasets. To sum up, we obtain the following conclusions: (i)
By investigating feature interactions, PNN gains better capacity on multi-field categorical data. (ii) Being both efficient and effective, PNN outperforms major state-of-the-art models. (iii)
Analogous to ``AND''/``OR'' gates, the product/add operations in PNN
provide a potential strategy for data representation, more specifically, rule representation.

In the future work, we will explore PNN with more general and complicated product layers. Besides, we are interested in explaining and visualizing the feature vectors learned by our models. We will investigate their properties, and further apply these node representations to other tasks.



\bibliographystyle{IEEEtran}
\bibliography{main}

\end{document}